\documentclass[12pt,a4paper,twoside]{article}

\usepackage[utf8]{inputenc}
\usepackage{mathptmx}
\usepackage[scaled=.90]{helvet}
\usepackage{courier}
\usepackage{tipa}

\usepackage[USenglish]{babel}
\usepackage[alsoload=abbr]{siunitx}

\usepackage{essv}
\usepackage{bibgerm}
\usepackage{fancyhdr}
\fancypagestyle{plain}{%
  \fancyhf{}
  \fancyfoot[LE,RO]{\thepage}

}
\usepackage{graphicx}
\usepackage{subfig}
\usepackage{authoraftertitle}
\title{%
  Progress in animation of an EMA-controlled\\
  tongue model for acoustic-visual speech synthesis
}
\author{%
  Ingmar Steiner, Slim Ouni\\[6pt]
  LORIA Speech Group\\
  Bat.~C,
  615 Rue du Jardin Botanique,
  54600 Villers-lès-Nancy,
  France
  \email{Firstname.Lastname@loria.fr}
}

\usepackage[numbers, sort&compress]{natbib}

\usepackage{hyperref}
\hypersetup{colorlinks, linkcolor=black, urlcolor=black, citecolor=black, pdfinfo={
  Title={\MyTitle},
  Author={\MyAuthor},
}}
\usepackage[all]{hypcap}

\usepackage[acronym, shortcuts, nowarn]{glossaries}
\glsdisablehyper
\newacronym{av}{AV}{acoustic-visual}
\newacronym{bb}{B-bone}{bézier bone}
\newacronym{ema}{EMA}{electromagnetic articulography}
\newacronym{gpu}{GPU}{graphics processing unit}
\newacronym{gui}{GUI}{graphical user interface}
\newacronym{ik}{IK}{inverse kinematics}
\newacronym{mri}{MRI}{magnetic resonance imaging}
\newacronym{nla}{NLA}{non-linear animation}
\newacronym{tts}{TTS}{text-to-speech}
\makeglossaries

\begin{document}
\maketitle

\begin{abstract}
We present a technique for the animation of a 3D kinematic tongue model, one component of the talking head of an \ac{av} speech synthesizer. The skeletal animation approach is adapted to make use of a deformable rig controlled by tongue motion capture data obtained with \ac{ema}, while the tongue surface is extracted from volumetric \ac{mri} data. Initial results are shown and future work outlined.

\end{abstract}


\glsresetall
\section{Introduction}

As part of ongoing research in developing a fully data-driven \ac{av} \ac{tts} synthesizer \cite{Toutios2010AVSP}, we integrate a tongue model to increase visual intelligibility and naturalness. To extend the kinematic paradigm used for facial animation in the synthesizer to tongue animation, we adapt state-of-the-art techniques of animation with motion-capture data for use with \ac{ema}.

\label{sec:visac}

Our \ac{av} synthesizer\footnote{\url{http://visac.loria.fr/}} is based on a non-uniform unit-selection \ac{tts} system for French \cite{Colotte2005IS}, concatenating bimodal units of acoustic and visual data, and extending the selection algorithm with visual target and join costs \cite{Musti2011AVSP}. The result is an application whose \ac{gui} features a ``talking head'' (i.e.\ computer-generated face), which is animated synchronously with the synthesized acoustic output.

This synthesizer depends on a speech corpus acquired by tracking marker points painted onto the face of a human speaker, using a stereoscopic high-speed camera array, with simultaneously recorded audio. While the acoustic data is used for waveform concatenation in a conventional unit-selection paradigm, the visual data is post-processed to obtain a dense, animated 3D point cloud representing the speaker's face. The points are interpreted as the vertices of a mesh, which is then rendered as an animated surface to generate the face of the talking head using a standard \emph{vertex animation} paradigm.

Due to the nature of the acquisition setup, no intra-oral articulatory motion data can be simultaneously captured. At the very least, any invasive instrumentation, such as \ac{ema} wires or transducer coils, would have a negative effect on the speaker's articulation and hence, the quality of the recorded audio; additional practical issues (e.g.\ coil detachment) would limit the length of the recording session, and by extension, the size of the speech corpus. As a consequence, the synthesizer's talking head currently features neither tongue nor teeth, which significantly decreases both the naturalness of its appearance and its visual intelligibility.

To address this shortcoming, we develop an independently animated 3D tongue and teeth model, which will be integrated into the talking head and eventually controlled by interfacing directly with the \ac{tts} synthesizer.


\section{\acs{ema}-based tongue model animation}
\glsreset{ema}

To maintain the data-driven paradigm of the \ac{av} synthesizer, the tongue model\footnote{For reasons of brevity, in the remainder of this paper, we will refer only to a \emph{tongue} model, but it should be noted that that such a model can easily encompass upper and lower teeth in addition to the tongue.} consists of a geometric mesh rendered in the \ac{gui} along with (or rather, ``behind'') the face. Since the primary purpose of the tongue model is to improve the \emph{visual} aspects of the synthesizer and it has no influence on the acoustics, there is no requirement for a complex tongue model to calculate the vocal tract transfer function, etc. Therefore, in contrast to previous work \cite[e.g.][]{Pelachaud1994CompAnim, Engwall2003SpeCom, King2005TVCG, Gerard2006SP, Vogt2006ISSP, Lu2009IS}, most of which attempts to predict tongue shape and/or motion by simulating the dynamics in one form or another, we must merely generate realistic tongue \emph{kinematics}, without having to model the anatomical structure of the human tongue or satisfy physical or biomechanical constraints.

This scenario allows us to make use of standard animation techniques using motion capture data. Specifically, we apply \ac{ema} using a Carstens AG500\footnote{Carstens Medizinelektronik GmbH, \url{http://www.articulograph.de/}} to obtain high-speed (\SI{200}{\Hz}), 3D motion capture data of the tongue during speech \cite{Kaburagi2005JASA}.

While other modalities might be used to acquire the shape of the tongue while speaking, their respective drawbacks make them ill-suited to our needs. For example, ultrasound tongue imaging tends to require extensive processing to track the mid-sagittal tongue contour and does not usually capture the tongue tip, while real-time \ac{mri} has a very low temporal resolution, and is currently possible only in a single slice.\footnote{3D cine-\acs{mri} of the vocal tract \cite{Takemoto2006JASA}, while possible, is far from realistic for the compilation of a full speech corpus sufficient for \acs{tts}.}

\subsection{Tongue motion capture}

Conventional motion capture modalities (as widely used e.g.\ in the animation industry) normally employ a camera array to track optical markers attached to the face or body of a human actor, producing data in the form of a 3D point cloud sampled over time. For facial animation, these points (given sufficient density) can be directly used as vertices of a mesh representing the surface of the face; this is the vertex animation approach taken in the \ac{av} synthesizer (see \hyperref[sec:visac]{above}).

For articulated body animation, however, the 3D points are normally used as transformation targets for the rigid bones of a hierarchically structured (usually humanoid) skeleton model. Much like the strings controlling a marionette, the skeletal transformations are then applied to a virtual character by deforming its geometric mesh accordingly, a widely used technique known as \emph{skeletal animation}.

Since current \ac{ema} technology allows the tracking of no more than 12 transducer coils (usually significantly fewer on the tongue), the resulting data is too sparse for vertex animation of the tongue surface. For this reason, we adopt a skeletal animation approach, but without enforcing a rigid structure, since the human tongue contains no bones and is extremely deformable. This issue is addressed \hyperref[sec:animation]{below}.

One advantage of \ac{ema} lies in the fact that the data produced is a set of 3D \emph{vectors}, not points, as the AG500 tracks the orientation, as well as position, of each transducer coil. Thus, the rotational information supplements, and compensates to some degree for the sparseness of, the positional data. Technically, this corresponds to motion capture approaches such as \cite{Molet1999Human}, although the geometry is of course quite different for the tongue than for a humanoid skeleton.

\begin{figure}
\begin{minipage}[t]{.475\textwidth}
  \includegraphics[width=\textwidth]{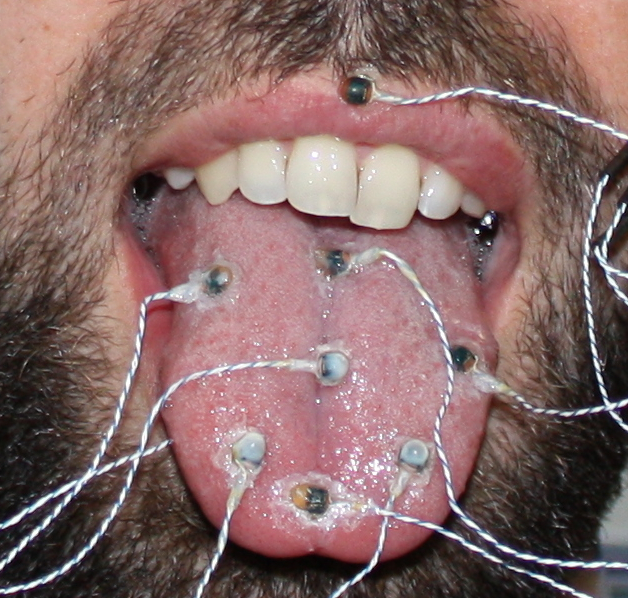}
  \caption{Coil layout for \acs{ema} test corpus}
  \label{fig:coils_photo}
\end{minipage}
\hfill
\begin{minipage}[t]{.475\textwidth}
  \includegraphics[width=\textwidth]{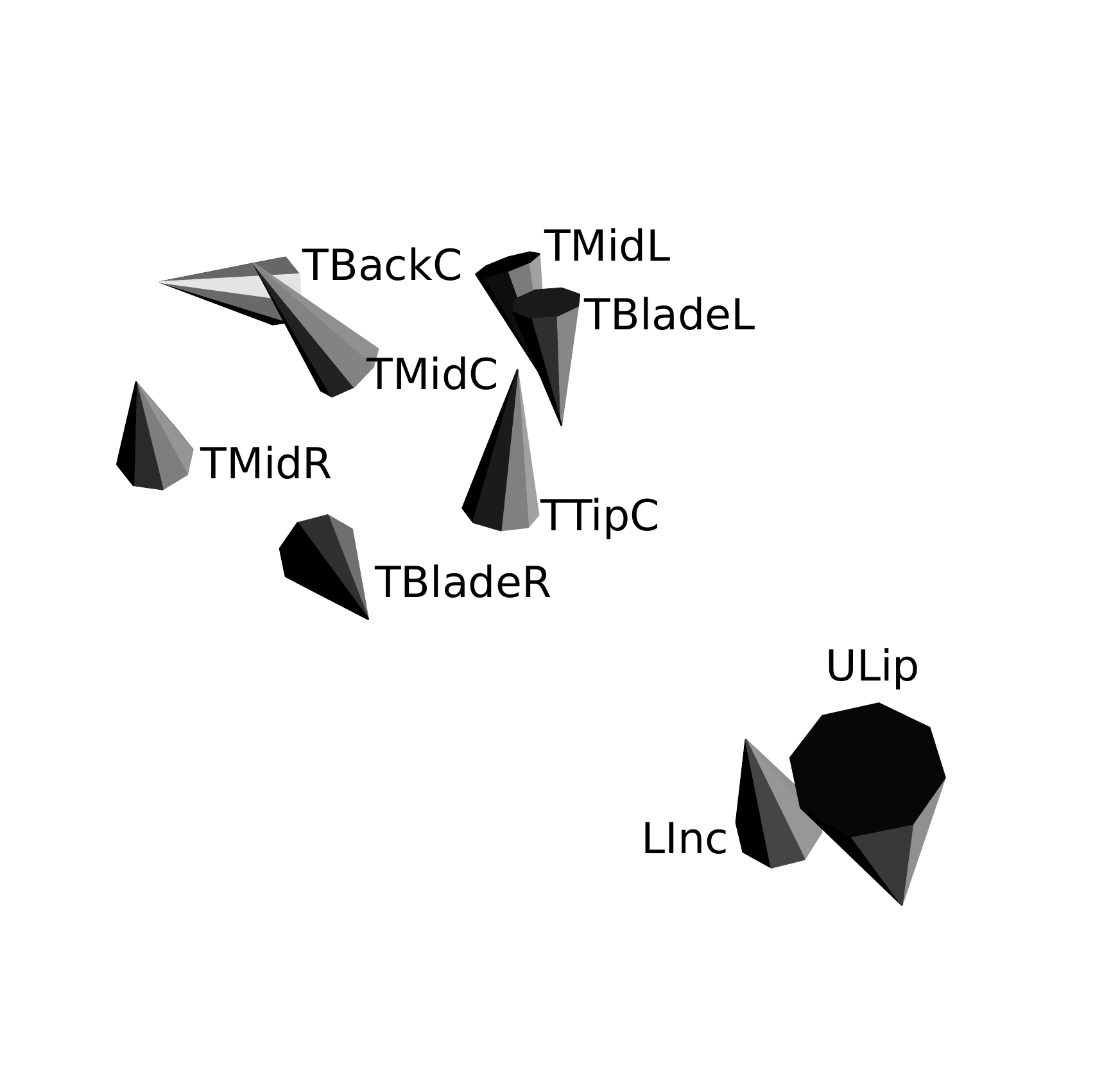}
  \caption{Perspective view of \acs{ema} coils rendered as primitive cones to visualize their orientation}
  \label{fig:coils_render}
\end{minipage}
\end{figure}

As a small \ac{ema} test corpus, we recorded one speaker using the AG500, with the following measurement coil layout: tongue tip center, tongue blade left/right, tongue mid center/left/right, tongue back center, lower incisor, upper lip (reference coils on bridge of nose and behind each ear). The exact arrangement can be seen in \autoref{fig:coils_photo}. The speech material comprises sustained vowels in the set \textipa{[i, y, u, e, \o, o, @, a]}, repetitive CV syllables permuting these vowels with the consonants in the set \textipa{[p, t, k, m, n, N, f, T, s, S, \c{c}, x, l, \textltilde]}, as well as 10 normal sentences in German and English, respectively. A 3D palate trace was also obtained.

We imported the raw \ac{ema} data as keyframes into the animation component of a fully-featured, open-source, 3D modeling and animation software suite,\footnote{Blender v2.5, \url{http://www.blender.org/}} using a custom plugin. Unlike point cloud based motion capture data contained in industry standard formats such as C3D \cite{Motion-Lab-Systems2008The-C3D-File}, this allows us to directly import the rotational data as well. As an example of the result, one frame is displayed in \autoref{fig:coils_render}. Within each frame of the animation, the \ac{ema} coil objects can provide the transformation targets for an arbitrary skeleton.

Once the motion capture data has been imported into the 3D software, it can be segmented into distinct actions for use and re-use in \ac{nla}. This allows us to manipulate and concatenate any number of frame sequences as atomic actions, and to synthesize new animations from them, using e.g.\ the 3D software's \ac{nla} editor (which, for these purposes, is conceptually similar to a gestural score in articulatory phonology \cite{Browman1992Phon}).

\subsection{Tongue model animation}
\label{sec:animation}

\begin{figure}
\subfloat[Perspective view of \acs{ema} coils (rendered as cones) and deformable skeletal rig in bind pose (\textipa{[a]} vowel). Tongue tip at center, oriented towards left; upper lip and lower incisor coils are visible further left. Adaptation struts (\acs{ik} targets, cf.\ text) are shown as thin rods connecting coils and \acsp{bb}]{\includegraphics[width=.475\textwidth]{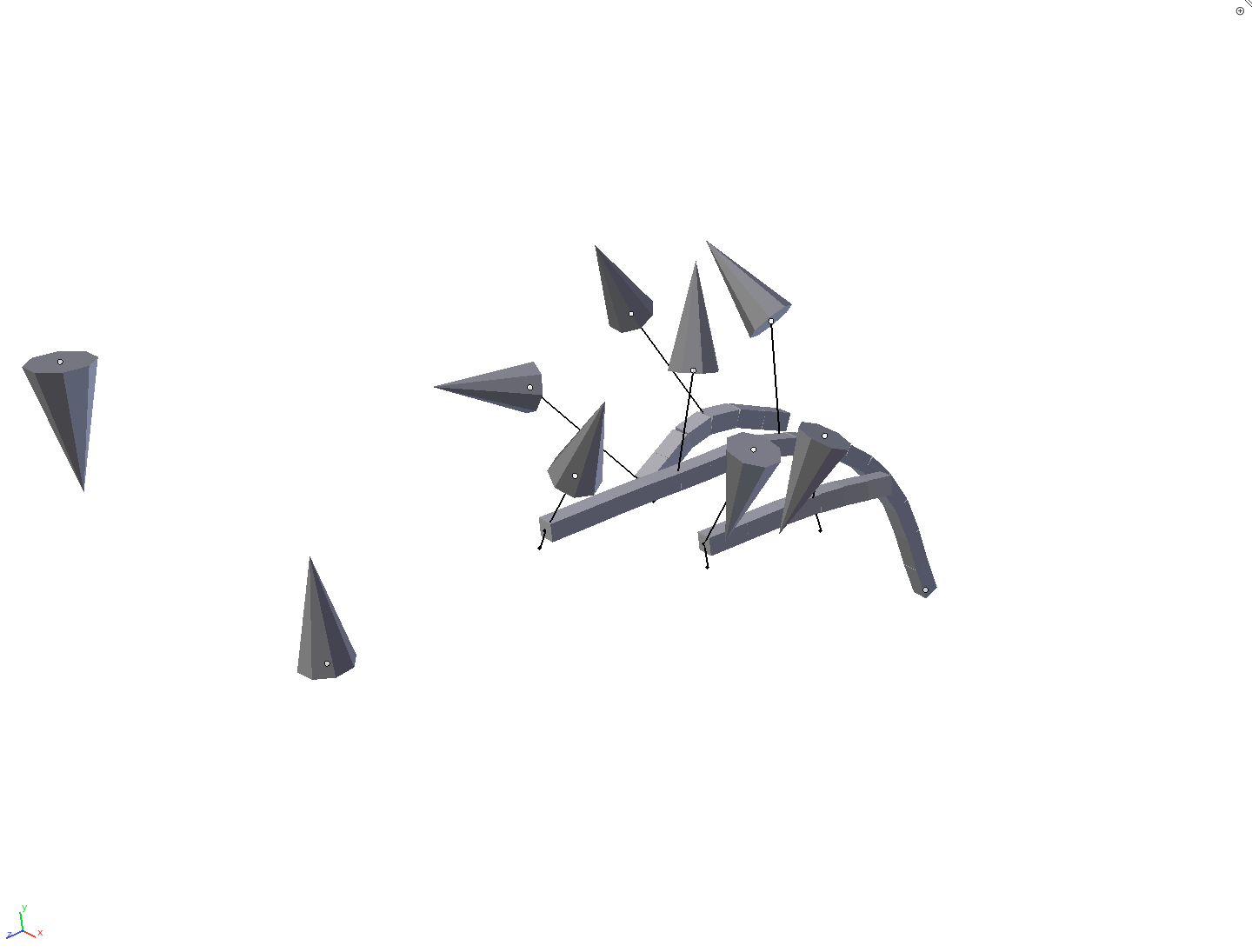}\label{fig:rig_a}}
\hfill
\subfloat[Like \ref{fig:rig_a}, but deformed according to \textipa{[t]} target pose]{\includegraphics[width=.475\textwidth]{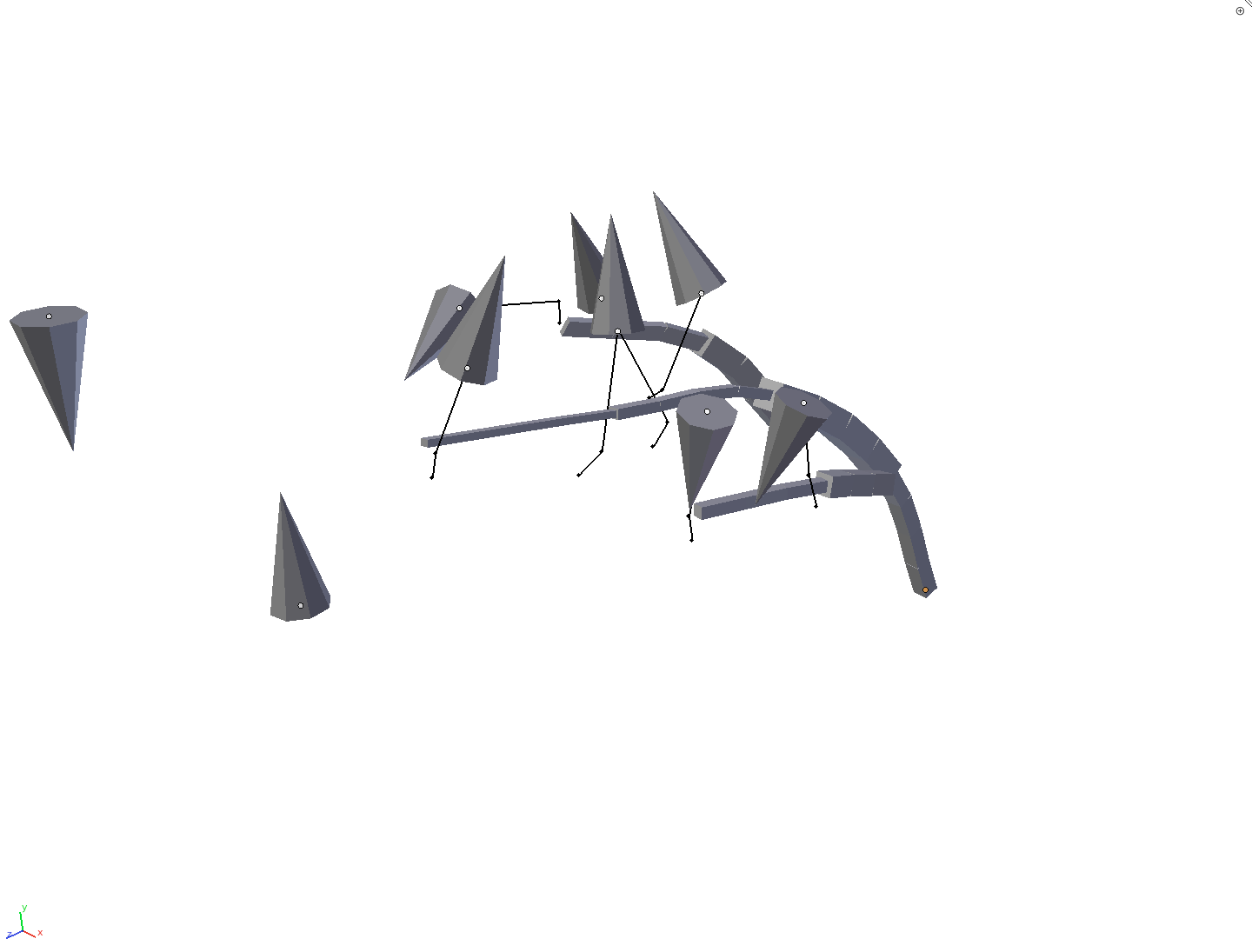}}
\\
\subfloat[Like \ref{fig:rig_a}, but with the tongue mesh bound to the rig. Its surface is color-shaded with a heat map visualizing the influence of the tongue-tip \acs{bb} on the mesh vertices (red$ = $full; blue$ = $none)]{\includegraphics[width=.475\textwidth]{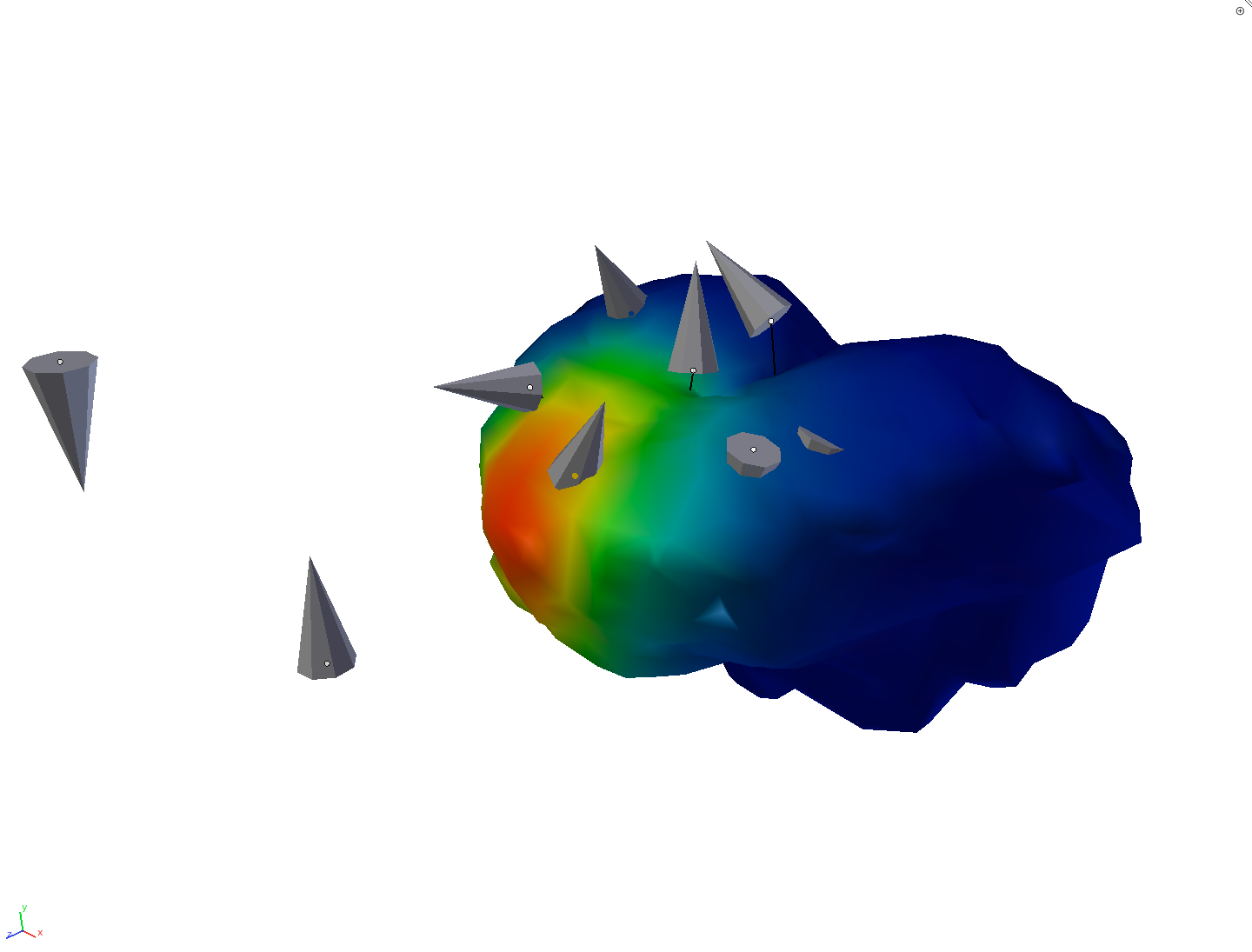}\label{fig:weight_a}}
\hfill
\subfloat[Like \ref{fig:weight_a}, but deformed according to \textipa{[t]} target pose]{\includegraphics[width=.475\textwidth]{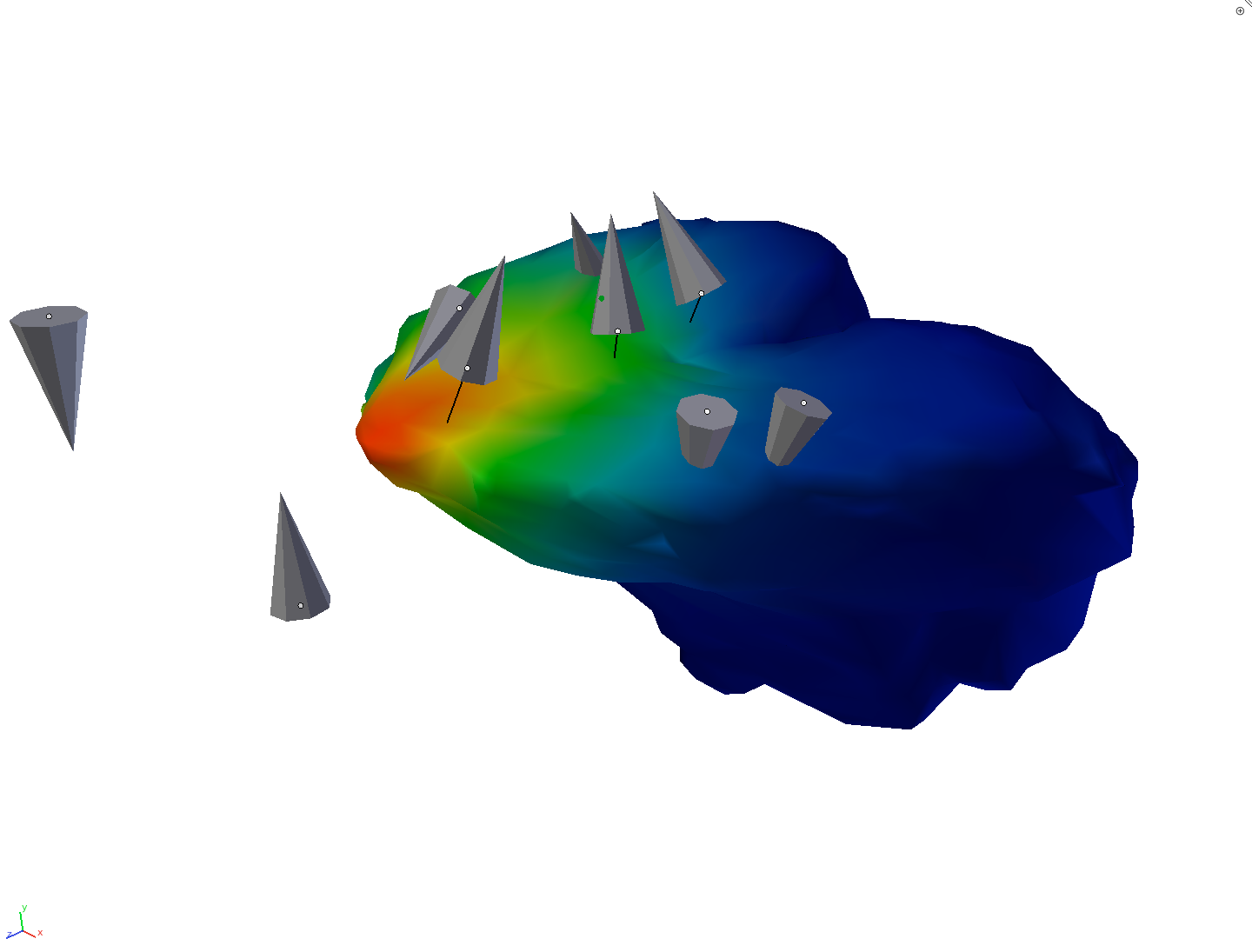}}
\\
\subfloat[Like \ref{fig:weight_a}, but showing only the tongue surface mesh, with the tip at center, oriented left]{\includegraphics[width=.475\textwidth]{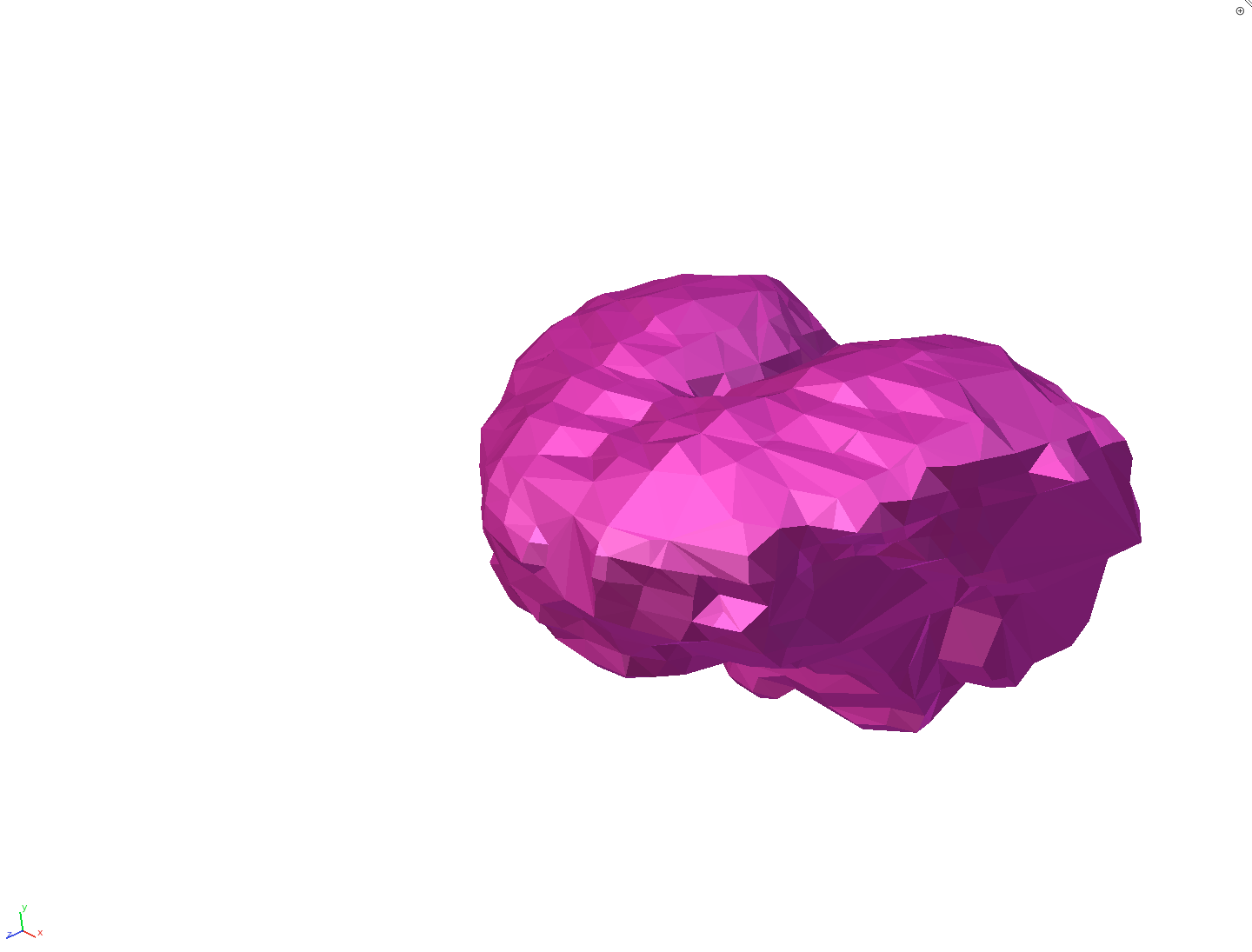}\label{fig:tongue_a}}
\hfill
\subfloat[Like \ref{fig:tongue_a}, but deformed according to \textipa{[t]} target pose]{\includegraphics[width=.475\textwidth]{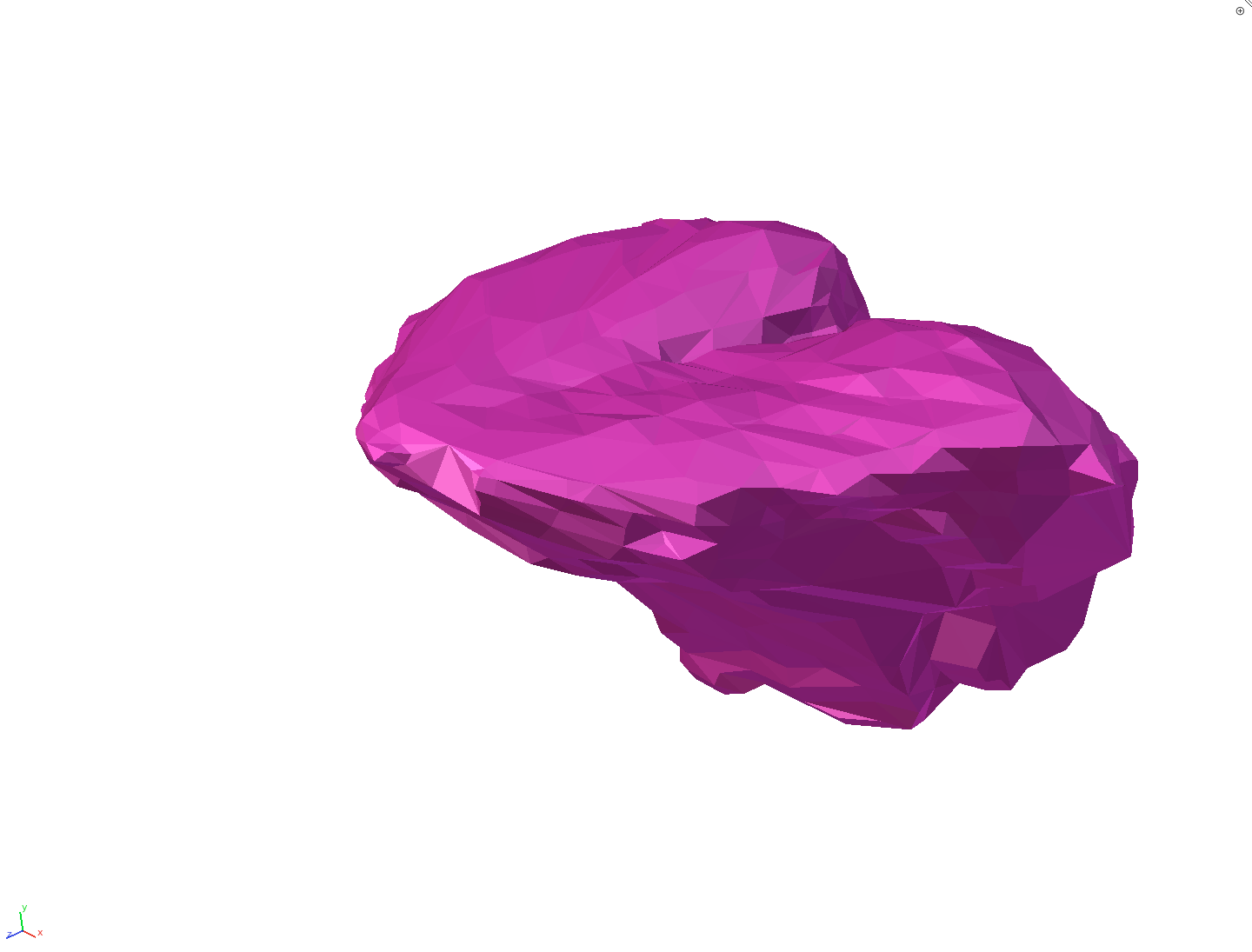}}
\caption{Tongue model in initial and final frame of one \textipa{[at]} cycle in the \acs{ema} test corpus}
\label{fig:riggedtongue}
\end{figure}

In order to use the tongue motion capture data to control a tongue model using skeletal animation, we design a simple skeleton as a rig for the tongue mesh. This rig consists of a central ``spine'', and two branches to allow (potentially asymmetric) lateral movement, such as grooving. Once again, it must be pointed out that this rig is unrelated to real tongue anatomy, although it could be argued that e.g.\ the spine corresponds roughly to the superior and inferior longitudinal muscles.

Of course, a skeleton of rigid bones is inadequate to mimic the flexibility of a real tongue. Our solution is to construct the rig using \emph{deformable} bones, so-called \acp{bb}, which can bend, twist, and stretch as required, governed by a set of constraining parameters.

The tongue model should be able to move independently of any specific \ac{ema} coil layout, since after all, the motion capture data represents \emph{observations} of tongue movements based on hidden dynamics. For this reason, and to maintain as much modularity and flexibility as possible in the design, the animation rig is not directly connected to the \ac{ema} coils in the motion capture data. Instead, we introduce an \emph{adaptation layer} in the form of ``struts'', each of which is connected to one coil, while the other end serves as a target for the rig's \acp{bb}. These struts can be adapted to any given \ac{ema} coil layout or rig structure.

With the struts in place and constrained to the movements of the \ac{ema} coils, the rig can be animated by using \emph{\ac{ik}} to determine the location, rotation, and deformation of each \ac{bb} for any given frame. The \ac{ik} are augmented by volume constraints, which inhibit potential ``bloating'' of the rig during \ac{bb} stretching.

The final component for tongue model animation is a mesh that represents the tongue surface, which is rendered in the \ac{gui} and deformed according to the skeletal animation. While this tongue mesh could be an arbitrary geometric structure, we use an isosurface extracted from a volumetric scan in a \ac{mri} speech corpus (from a different speaker; voxel size $\SI{1.09}{mm}\times\SI{1.09}{\mm}\times\SI{4}{\mm}$). The tongue in this scan was manually segmented using a graphics tablet and open-source medical imaging software.\footnote{OsiriX v3.9, \url{http://www.osirix-viewer.com/}}

The resulting tongue mesh was manually registered to the \ac{ema} coil positions in a neutral \emph{bind pose}. The skeletal rig was then embedded, and vertex groups in the tongue mesh assigned automatically to each \ac{bb}. As the motion capture data animates the rig using \ac{ik}, the tongue mesh is deformed accordingly, approximately following the \ac{ema} coils. \autoref{fig:riggedtongue} displays the initial and final frame in one cycle of repetitive \textipa{[ta]} articulation in the \ac{ema} test corpus. In an informal evaluation, our technique appears to produce satisfactory results, and encourages us to pursue and refine this approach to tongue model animation.

\section{Discussion and Outlook}

We have presented a technique to animate a kinematic tongue model, based on volumetric vocal tract \ac{mri} data, using skeletal animation with a flexible rig, controlled by motion capture data acquired with \ac{ema}, and implemented with off-the-shelf, open-source software. While this approach appears promising, it is still under development, and there are various issues which must be addressed before the tongue model can be integrated into our \ac{av} \ac{tts} synthesizer as intended.

\begin{itemize}
  \item Upper and lower teeth can be added to the model using the same data and animation technique, albeit with a conventional, rigid skeleton. These will then be rendered in the synthesizer's \ac{gui} along with the face and tongue.
  \item The tongue mesh used here is quite rough, and registration with the \ac{ema} data does not produce the best fit, owing to differences between the speakers' vocal tract geometries and articulatory target positions, quite possibly exacerbated by the effects of supine posture during \ac{mri} scanning \cite[e.g.][]{Kitamura2005AST}. A more suitable mesh might be obtained by scanning the tongue of the same speaker used for the \ac{ema} motion capture data, at a higher resolution.
  \item Registration of the tongue mesh into the 3D space of the tongue model should be performed in a partially or fully automatic way, using landmarks available in both \ac{mri} and \ac{ema} modalities \cite[cf.][]{Aron2009ICASSP}, such as the 3D palate trace and/or high-contrast markers at the positions of the reference coils.
  \item The reliability of \ac{ema} position and orientation data is sometimes unpredictable. This could be due to the algorithms used to process the raw amplitude data, faulty hardware, interference (even within the coil layout itself), or any combination of such factors. However, since any such errors are immediately visible in the animation of the tongue model by introducing implausible deformations, we are working on methods both to clean the \ac{ema} data itself, and to make the tongue model less susceptible to such outlier trajectory segments.
  \item To evaluate the performance of the animation technique, factors such as skin deformation and distance of \ac{ema} coils from the tongue model surface should be monitored. The structure of the skeletal rig can be independently refined, optimizing its ability to generate realistic tongue poses. Its embedding into the tongue mesh should preferably be performed using a robust automatic method \cite[e.g.][]{Baran2007SIGGRAPH}.
  \item The 3D palate trace can be used to add a palate surface mesh to the tongue model. For both the palate and the teeth, the model could also be augmented with automatic collision detection by accessing the 3D software's integrated physics engine.\footnote{Bullet physics library, \url{http://www.bulletphysics.com/}}
\end{itemize}

For an interactive application such as the \ac{av} synthesizer \ac{gui}, it is impractical to incur the performance overhead of an elaborate 3D rendering engine, especially when a non-trivial processing load is required for the bimodal unit-selection. Instead, we anticipate integrating the tongue model into the talking head using a more lightweight, real-time capable 3D game engine, which may even offload the visual computation to dedicated graphics hardware. The advantage of using keyframe-based, \ac{nla} actions is that they can be ported into such engines as animated 3D models, using common interchange formats.\footnote{For instance, COLLADA (\url{http://www.collada.org/}) or OGRE (\url{http://www.ogre3d.org/})} Although the skeletal rig could be accessed and manipulated directly, this ``pre-packaging'' of animation actions also avoids the complexity, or perhaps even unavailability, of advanced features such as \acp{bb} or \ac{ik} in those engines.

The final integration challenge is to interface the tongue model directly with the \ac{tts} system to synthesize the correct animation actions with appropriate timings. This task might be accomplished using a diphone synthesis style approach, or even action unit-selection, and will be addressed in the near future.

\section*{Acknowledgments}

We owe our thanks to Sébastien Demange for assistance during the recording of the \ac{ema} test corpus, and to Korin Richmond for providing the means to record the \ac{mri} data used here.

\bibliographystyle{gerabbrv}
\bibliography{ref/tongue}

\end{document}